\pdfoutput=1

\documentclass{article}
\usepackage{hyperref}
\newcommand{\showComments}{yes}
\usepackage{comment}
\usepackage{color}
\newcommand{\note}[2]{\ifthenelse{\equal{\showComments}{yes}}{\textcolor{#1}{#2}}{}}

\usepackage[utf8]{inputenc} 
\usepackage[T1]{fontenc}
\usepackage{url}
\usepackage{booktabs}       
\usepackage{amsfonts}       
\usepackage{nicefrac}       
\usepackage{microtype}      
\usepackage{ifthen}
\usepackage[linesnumbered,ruled]{algorithm2e}
\usepackage[cmex10]{amsmath}
\usepackage{amsthm, amssymb, bbm, bm, gensymb}
\usepackage{graphicx}
\usepackage{subfig}
\usepackage{wrapfig}
\usepackage[font=small]{caption}
\usepackage[colorinlistoftodos,prependcaption]{todonotes}
\usepackage{hyperref}
\usepackage{cleveref}

\usepackage{fullpage,authblk}

\newtheorem{prop}{Proposition}

\newcommand{\alldata}{\mathbf{X}}
\newcommand{\onedata}{\mathbf{x}}

\newcommand{\alltarget}{\mathbf{y}}
\newcommand{\onetarget}{y}

\newcommand{\randlatent}{\mathbf{Z}}
\newcommand{\onelatent}{\mathbf{z}}
\newcommand{\latentspace}{\mathcal{Z}}

\newcommand{\kernel}{K}

\newcommand{\smallkernel}{k}
\newcommand{\wtkernel}{\kappa}

\newcommand{\latfunc}{g}

\newcommand{\probfunc}{p}
\newcommand{\probfuncspace}{\mathcal{P}}

\newcommand{\numtrain}{n}

\newcommand{\highdim}{D}
\newcommand{\lowdim}{d}

\newcommand{\negll}{L}

\newcommand{\weights}{\mathbf{w}}
\newcommand{\randweights}{\mathbf{W}}
\newcommand{\tfweights}{\mathbf{u}}

\newcommand{\numweights}{m}

\newcommand{\ncaprob}{q}

\newcommand{\rkhs}{\mathcal{H}}
\newcommand{\tf}{T}
\newcommand{\shift}{s}

\newcommand{\classind}{c}

\newcommand{\altshift}{r}

\graphicspath{{Figures/}}

\DeclareMathOperator*{\argmax}{arg\,max}
\DeclareMathOperator*{\argmin}{arg\,min}

\crefname{equation}{}{}
\Crefname{equation}{}{}
\crefname{thm}{theorem}{theorems}
\Crefname{thm}{Theorem}{Theorems}
\crefname{clm}{claim}{claims}
\Crefname{clm}{Claim}{Claims}
\Crefname{coro}{Corollary}{Corollaries}
\Crefname{lem}{Lemma}{Lemmas}
\Crefname{sec}{Section}{Sections}
\crefname{app}{appendix}{appendices}
\Crefname{app}{Appendix}{Appendices}
\crefname{prop}{proposition}{propositions}
\Crefname{prop}{Proposition}{Propositions}
\Crefname{propty}{Property}{Properties}
\crefname{figure}{fig.}{figures}
\Crefname{figure}{Fig.}{Figures}
\crefname{defn}{definition}{definitions}
\Crefname{defn}{Definition}{Definitions}
\crefname{fact}{fact}{facts}
\Crefname{fact}{Fact}{Facts}
\crefname{appendix}{appendix}{appendices}
\Crefname{appendix}{Appendix}{Appendices}
\crefname{algo}{algorithm}{algorithms}
\Crefname{algo}{Algorithm}{Algorithms}
\crefname{algorithm}{algorithm}{algorithms}
\Crefname{algorithm}{Algorithm}{Algorithms}
\crefname{tbl}{table}{table}
\Crefname{tbl}{Table}{Table}
\crefname{table}{table}{table}
\Crefname{table}{Table}{Table}
\crefname{algorithm}{algorithm}{algorithms}
\Crefname{algorithm}{Algorithm}{Algorithms}

\crefname{conj}{conjecture}{conjectures}
\Crefname{conj}{Conjecture}{Conjectures}
\crefname{obs}{observation}{observations}
\Crefname{obs}{Observation}{Observations}

\title{Probabilistic Neighbourhood Component Analysis: \\Sample Efficient Uncertainty Estimation in Deep Learning}

\author[1]{Ankur Mallick}
\author[1]{Chaitanya Dwivedi}
\author[2]{Bhavya Kailkhura}
\author[1]{Gauri Joshi}
\author[2]{T. Yong-Jin Han}
\affil[1]{Carnegie Mellon University}
\affil[2]{Lawrence Livermore National Laboratory}

\begin{document}

\maketitle

\begin{abstract}
While Deep Neural Networks (DNNs) achieve state-of-the-art accuracy in various applications, they often fall short in accurately estimating their predictive uncertainty and, in turn, fail to recognize when these predictions may be wrong. 
Several uncertainty-aware models, such as Bayesian Neural Network (BNNs) and Deep Ensembles have been proposed in the literature for quantifying predictive uncertainty. However, research in this area has been largely confined to the big data regime.
In this work, we show that the uncertainty estimation capability of state-of-the-art BNNs and Deep Ensemble models degrades significantly when the amount of training data is small. 
To address the issue of accurate uncertainty estimation in the small-data regime, we propose a probabilistic generalization of the popular sample-efficient non-parametric kNN approach. Our approach enables deep kNN classifier to accurately quantify underlying uncertainties in its prediction. We demonstrate the usefulness of the proposed approach by achieving superior uncertainty quantification as compared to state-of-the-art on a real-world application of COVID-19 diagnosis from chest X-Rays. Our code is available at \href{https://github.com/ankurmallick/sample-efficient-uq}{https://github.com/ankurmallick/sample-efficient-uq}.
\end{abstract}

\section{Introduction}
Deep Neural Networks (DNNs) have achieved remarkable success in a wide range of applications where a large amount of labeled training data is available \cite{krizhevsky2012imagenet,graves2013speech}. However, in many emerging applications of machine learning such as diagnosis and treatment of novel coronavirus disease (COVID-19) \cite{cohen2020covid} a large labeled training datasets may not be available. Furthermore, test data in these applications may deviate from the training data distribution, e.g., due to sample selection bias, nonstationarity, and even can be from Out-of-Distribution in some extreme cases~\cite{bulusu2020anomalous}. Note that several of these applications are high-regret in nature implying that incorrect decisions or predictions have significant costs. Therefore, such applications require not only achieving high accuracy but also accurate quantification of predictive uncertainties. Accurate predictive uncertainty in these applications can help practitioners to assess the true performance and risks and to decide whether the model predictions should (or should not) be trusted.

\begin{figure}[!t]
    \centering
    \includegraphics[width=0.95\linewidth]{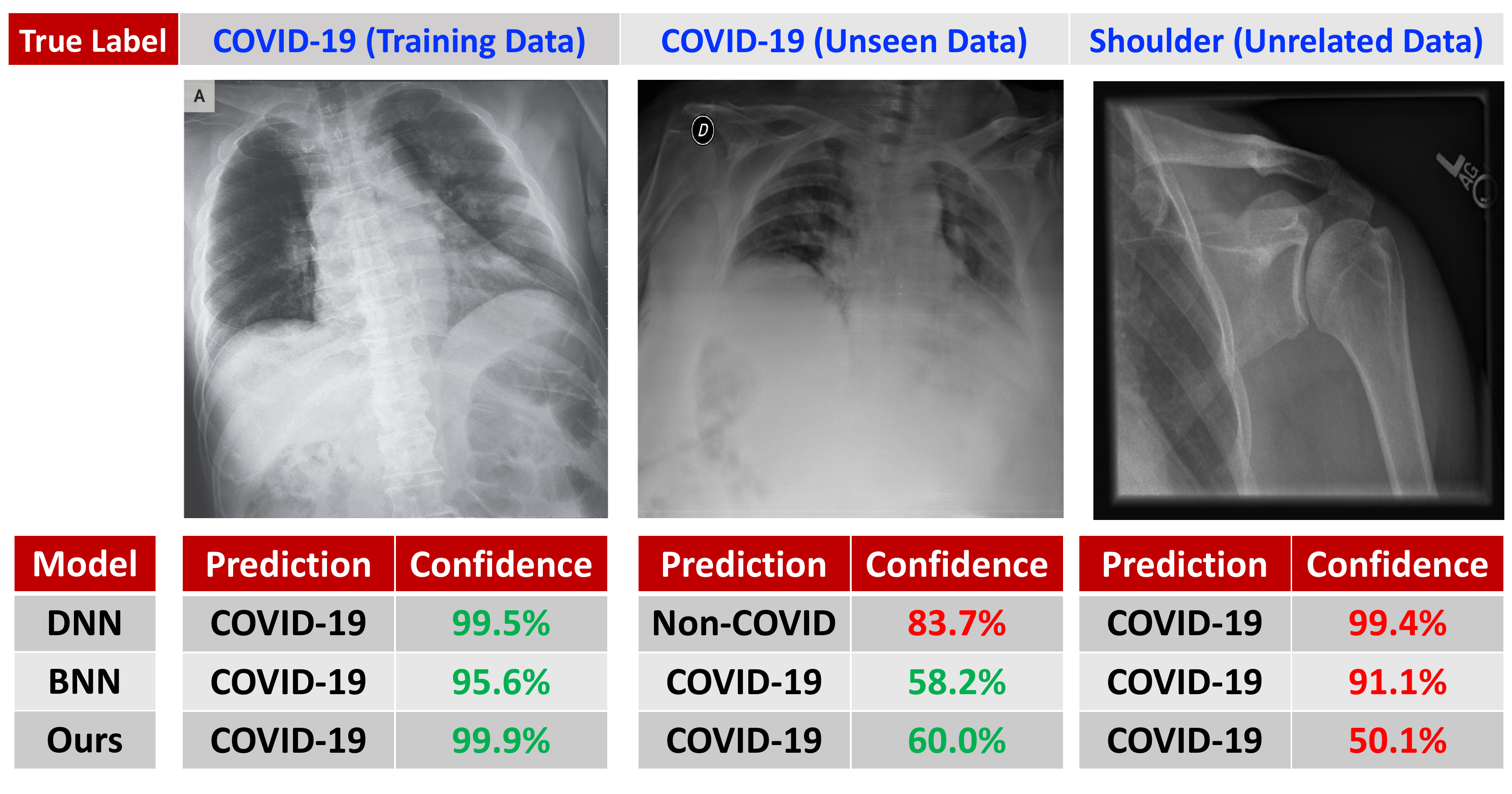}
    \vspace{-0.15in}
    \caption{The predictions of deep learning models that are trained to detect the presence of COVID-19 in chest X-Ray images from~\cite{cohen2020covid}. (a) All models correctly classify in-distribution data from \cite{cohen2020covid}, (b) Uncertainty-aware models (BNN, and our model PNCA) perform better than DNN when test data is from a different source~\cite{ActualMed}, (c) As opposed to proposed PNCA, both BNN and DNN make overconfident misclassification on Out-of-Distribution data~\cite{rajpurkar2017mura} (e.g., classifying shoulder X-Ray as COVID-19).
    }
    \label{fig:Teaser}
\end{figure}

Unfortunately, DNNs often make overconfident predictions in the presence of distributional shifts and Out-of-Distribution data.
As an example, \Cref{fig:Teaser} shows the predictions of different deep learning models trained to detect the presence of COVID-19 from chest X-ray images. All models achieve similar accuracy ($\sim 80\%$) on in-distribution validation data. However, their quality of uncertainties is widely varied as explained next.
While all models are forced to output some prediction, on every input image, we would want a model to not be very confident on input data that is very different from the data used to train it. However, we observe that state-of-the-art deep learning models make highly overconfident predictions on Out-of-Distribution data \cite{rajpurkar2017mura}.
Interestingly, we found that even popular uncertainty-aware models, (e.g., BNNs, deep ensembles) that are designed to address this precise issue, perform poorly in small data regime. 
This is an extremely problematic issue especially owing to the flurry of papers that have been attempting to use DNNs for detecting COVID-19 using chest X-Ray images \cite{minaee2020deep, cohen2020predicting, wang2020covid} as real-world test data is almost always different as compared to the training data.

While there have been separate efforts on improving the sample efficiency~\cite{mallick2019deep} and accurate uncertainty estimation~\cite{gal2016uncertainty} of deep learning, to the best of our knowledge there has not been any effort on studying these seemingly different issues in a unified manner. Therefore, this paper takes some initial steps towards (a) studying the effect of training data on the quality of uncertainty and (b) developing sample efficient uncertainty-aware predictive models. 
Specifically, to overcome the challenge of providing accurate uncertainties without compromising the accuracy in the small-data regime, we propose a probabilistic generalization of the popular non-parametric kNN approach (referred to as probabilistic neighborhood component analysis (PNCA)). 
By mapping data into distributions in a latent space before performing classification, we enable a deep kNN classifier to accurately quantify underlying uncertainties in its prediction. Following~\cite{lakshminarayanan2017simple, snoek2019can}, for a meaningful and effective performance evaluation, we compare the quality of predictive uncertainty of different models under conditions of distributional shift and Out-of-Distribution. We empirically show that the proposed PNCA approach achieves significantly better uncertainty estimation performance as compared to state-of-the-art approaches in small data regime. 
\section{Probabilistic Neighbourhood Component Analysis}
\label{sec:PNCA}
In this section, we describe our model to achieve sample-efficient and uncertainty-aware classification. The details of the algorithm and proof of Proposition 1 are presented in \Cref{app:Proof}.
\subsection{Neighbourhood Components Analysis (NCA)}
Our approach is a generalization of NCA proposed in \cite{goldberger2005neighbourhood} wherein the authors learn a distance metric for kNN classification of points $\onedata_i, \ldots, \onedata_\numtrain \in \mathbb{R}^{\highdim}$ with corresponding class labels $\onetarget_1,\ldots,\onetarget_n$. A data point $\onedata$ is projected into a latent space $\latentspace \subseteq \mathbb{R}^{\lowdim}$ to give an embedding $\onelatent=\latfunc_\weights(\onedata)$. Here $\latfunc$ can be a linear transformation like a  $\lowdim \times \highdim$ matrix or a non-linear transformation like a neural network with a $\lowdim-$dimensional output, and $\weights$ are the parameters of the transformation. The probability of a point $\onedata_i$ selecting another point $\onedata_j$ as its neighbor is given by applying a softmax activation to the distance between points in the latent space
\begin{align}
    \ncaprob_{ij} = \frac{\exp(-||\onelatent_i - \onelatent_j||^2)}{\sum_{i' \neq i}\exp(-||\onelatent_i - \onelatent_{i'}||^2)}, \quad \ncaprob_{ii} = 0.  \label{eq:NCA_probs}
\end{align}
The probability of $\onedata_i$ selecting a point in the same class as itself is given by $\ncaprob_i = \sum_{j:\onetarget_j=\onetarget_i}\ncaprob_{ij}$ and the optimal model parameters are obtained by minimizing the loss 
\begin{align}
\negll(\weights) = -\sum_{i}\log(\ncaprob_i), \label{eq:NCA}
\end{align}
which is the negative log-likelihood of the data under our the model. The authors of \cite{goldberger2005neighbourhood} experiment with a variety of transformations $\latfunc_\weights(.)$ and classification tasks and show that NCA achieves competitive accuracy.
\subsection{Our Model} 
The lack of data may cause the NCA model to overfit when learning the weights by optimizing the loss in \Cref{eq:NCA}. We expect that the uncertainty due to the scarcity of training data can be better captured by \emph{probability distributions} in the latent space $\latentspace$ than by individual data samples. Therefore, we propose a probabilistic generalization of the model, PNCA, which learns a distribution over the model parameters $\weights$ and, thus, deals with both the lack of training data and the task of accurate uncertainty estimation.

\textbf{Latent Space Mapping using Probabilistic Neural Networks.} Each data point $\onedata$ passes through a probabilistic neural network with parameters $\randweights \sim \probfunc(\randweights)$ to give a random variable $\randlatent = \latfunc_\randweights(\onedata) \in \latentspace$. Due to the stochasticity of $\randweights$, each data point $\onedata_i$ corresponds to a different distribution $\probfunc(\randlatent|\onedata_i)$ in the latent space. 

\textbf{NCA over Latent Distributions.} Observe that the individual terms in the softmax activation correspond to a \emph{kernel} between latent embeddings $\onelatent$, e.g., the squared exponential kernel $\smallkernel(\onelatent,\onelatent') = \exp(-||\onelatent-\onelatent'||^{2})$. Since in our approach the embedding corresponding to a data point $\onedata_i$ is the probability distribution $\probfunc(\randlatent|\onedata_i)$, we propose to use the following kernel between distributions
\begin{align}
    \kernel_{ij} = \mathbb{E}_{\onelatent \sim \probfunc(\randlatent|\onedata_i),\ \onelatent' \sim \probfunc(\randlatent|\onedata_j)}[\smallkernel(\onelatent,\onelatent')], \label{eq:kernij}
\end{align}
where $\kernel_{ij}$ corresponds to the inner product between distributions $\probfunc(\randlatent|\onedata_i)$ and $\probfunc(\randlatent|\onedata_j)$ in the Reproducing Kernel Hilbert Space (RKHS) $\rkhs_\smallkernel$ defined by the kernel $\smallkernel$ \cite{muandet2017kernel} and, thus, captures similarity between distributions in the same way as $\smallkernel(\onelatent,\onelatent')$ captures the similarity between individual embeddings in NCA.
\begin{figure*}[t]
  \centering
    \subfloat[Accuracy (Rotated-MNIST)]{\includegraphics[width=0.33\linewidth]{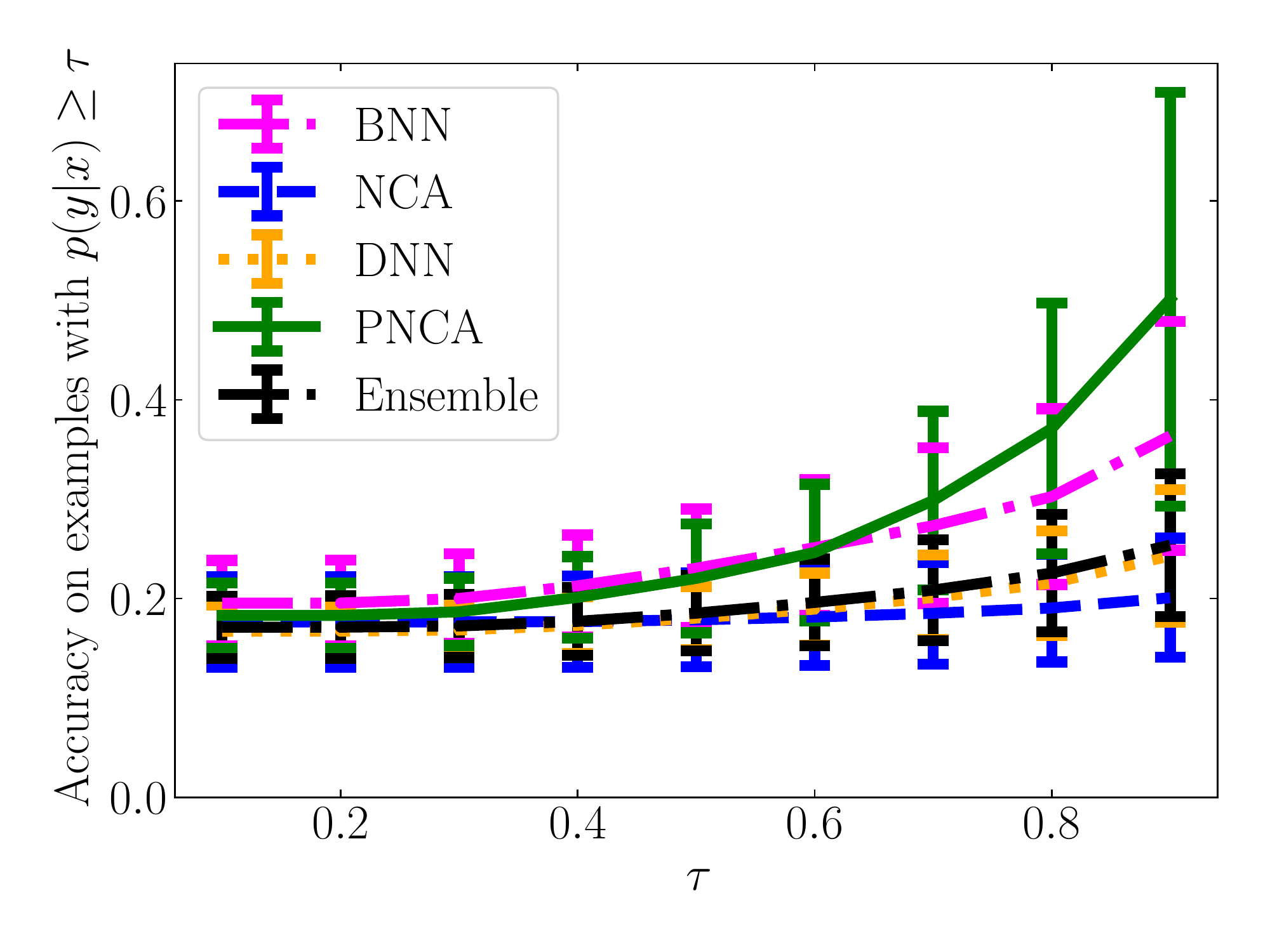}\label{fig:MNIST_rot_acc}}
   \subfloat[Count (Rotated-MNIST)]{\includegraphics[width=0.33\linewidth]{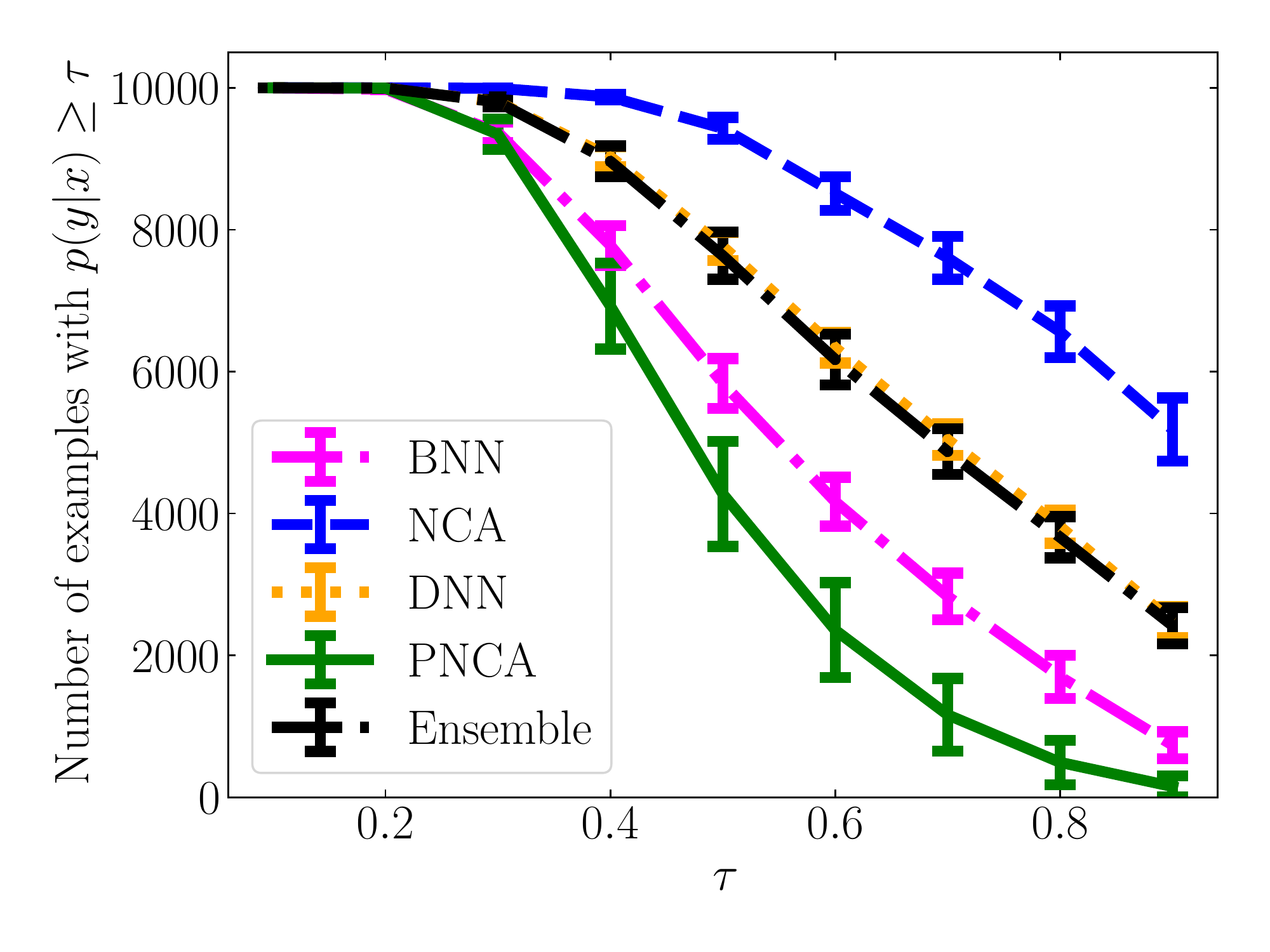}\label{fig:MNIST_rot_count}}
   \subfloat[Confidence on OoD (Not-MNIST)]{\includegraphics[width=0.33\linewidth]{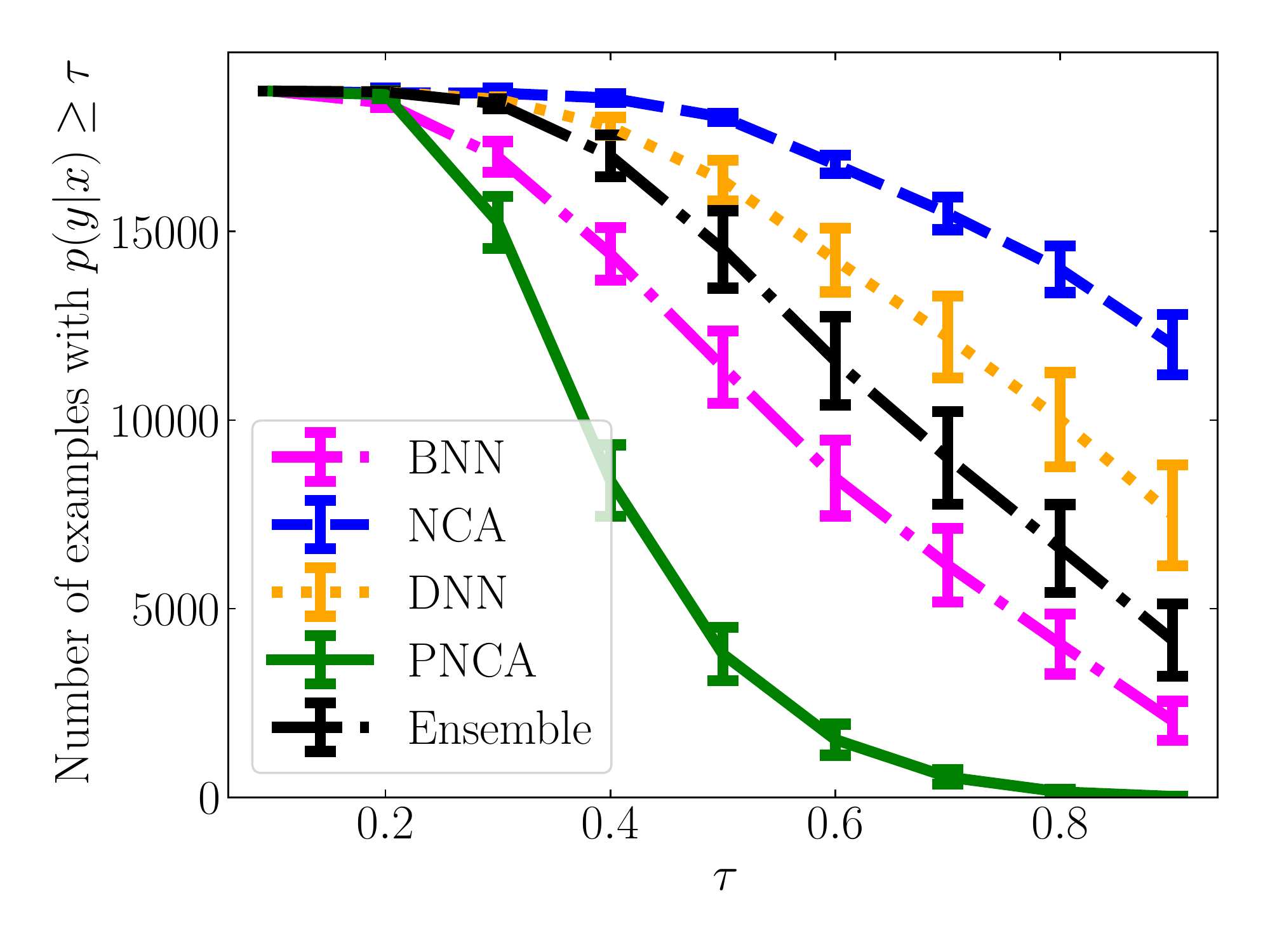}\label{fig:MNIST_OOD_count}}
   \caption{Results on MNIST: (a) PNCA has the largest fraction of correctly classified examples in the high-confidence region for Rotated MNIST. (b) and (c) PNCA generally has much lower confidence, i.e., better Uncertainty Quantification (UQ) than rest of the models on Out-of-Distribution Data -- both Rotated-MNIST and not-MNIST. See Fig. \ref{fig:MNIST_app} for a comparison by varying number of training samples.}
  \label{fig:MNIST}
\end{figure*}
\subsection{Training Algorithm} 
The forward pass described above, is used to compute a kernel between data points $\onedata_i$ and $\onedata_j$ which can then be used to compute the probability of $\onedata_i$ selecting $\onedata_j$ in an analogous fashion to NCA as
\begin{align}
    \ncaprob_{ij} = {\kernel_{ij}}/{\sum_{i' \neq i}\kernel_{ii'}}, \quad \ncaprob_{ii} = 0. \label{eq:PNCA_probs}
\end{align}
Since the latent embedding $\randlatent$ for a data point $\onedata$ is given by $\randlatent=\latfunc_{\randweights}(\onedata)$, $\randweights \sim \probfunc(\randweights)$, we can rewrite \Cref{eq:kernij} as 
\begin{align}
\kernel_{ij} &= \mathbb{E}_{\weights,\weights' \sim \probfunc(\randweights)}[\smallkernel(\latfunc_{\weights}(\onedata_i),\latfunc_{\weights'}(\onedata_j))] = \kernel_{ij}[\probfunc]. \label{eq:wtkernij}
\end{align}
Thus, we can view $\kernel_{ij}$ as a \emph{functional} of $\probfunc(\randweights)$. The negative log likelihood in \Cref{eq:NCA} is then also a functional of $\probfunc$. The optimal distribution over the model parameters can be obtained by solving 
\begin{align}
    \probfunc^{*}(\randweights) = \argmin_{\probfunc(\randweights) \in \probfuncspace} \negll[\probfunc]. \label{eq:PNCA}
\end{align}
The choice of $\probfuncspace$ is critical to the success of this approach.
Following \cite{liu2016stein}, we choose $\probfuncspace$ to be $\probfuncspace = \{\probfunc(\tfweights)|\tfweights = \weights + \shift(\weights), \weights \sim \probfunc_{0}(\weights), \shift \in \rkhs_\wtkernel\}$ where  $\rkhs_\wtkernel$ is a RKHS given by a kernel $\wtkernel$ between model parameters $\weights$ (note that this is \emph{different} from the RKHS $\rkhs_\smallkernel$ into which distributions in the latent space $\latentspace$ are embedded and which is given by the kernel $\smallkernel$). This choice of $\probfuncspace$ includes all smooth transformations from the initial distribution $\probfunc_{0}$, and the optimization problem $\Cref{eq:PNCA}$ now reduces to computing the optimal shift $\shift^{*}(\weights)$. 
Next, we provide an expression for the functional gradient of the negative log-likelihood under our model with respect to the shift $\shift$.
\begin{prop}
If we draw $\numweights$ realizations of model parameters $\weights_1,\ldots,\weights_\numweights \sim \probfunc(\weights)$, $\probfunc \in \probfuncspace$, then
\begin{align}
    \nabla_{\shift}\negll\mid_{\shift = 0} &\simeq \sum_{l = 1}^{\numweights}\wtkernel(\weights_l,.)\nabla_{\weights_l}\hat{\negll}(\weights_1,\ldots,\weights_\numweights)\label{eq:empfingrad}
\end{align}
where $\hat{\negll}$ is given by substituting $\hat{\kernel}_{ij} = \frac{1}{m^2}\sum_{l,l'}\smallkernel(\latfunc_{\weights_{l}}(\onedata_i),\latfunc_{\weights_{l'}}(\onedata_j))$ in \Cref{eq:NCA}.
\end{prop}
To estimate the optimal shift, $\shift^{*}$ (or optimal distribution $\probfunc^{*}$) we draw an initial set of parameters $\weights_i \sim \probfunc_{0}(\weights)$ and iteratively apply the functional gradient descent transformation $\tfweights = \weights - \epsilon\nabla_{\shift}\negll\mid_{\shift=0}$ as described in \Cref{algo:PNCA} in \Cref{app:Proof}

\section{Experiments}
\begin{figure*}[ht]
  \centering
    \subfloat[Accuracy (COVID-V2)]{\includegraphics[width=0.33\linewidth]{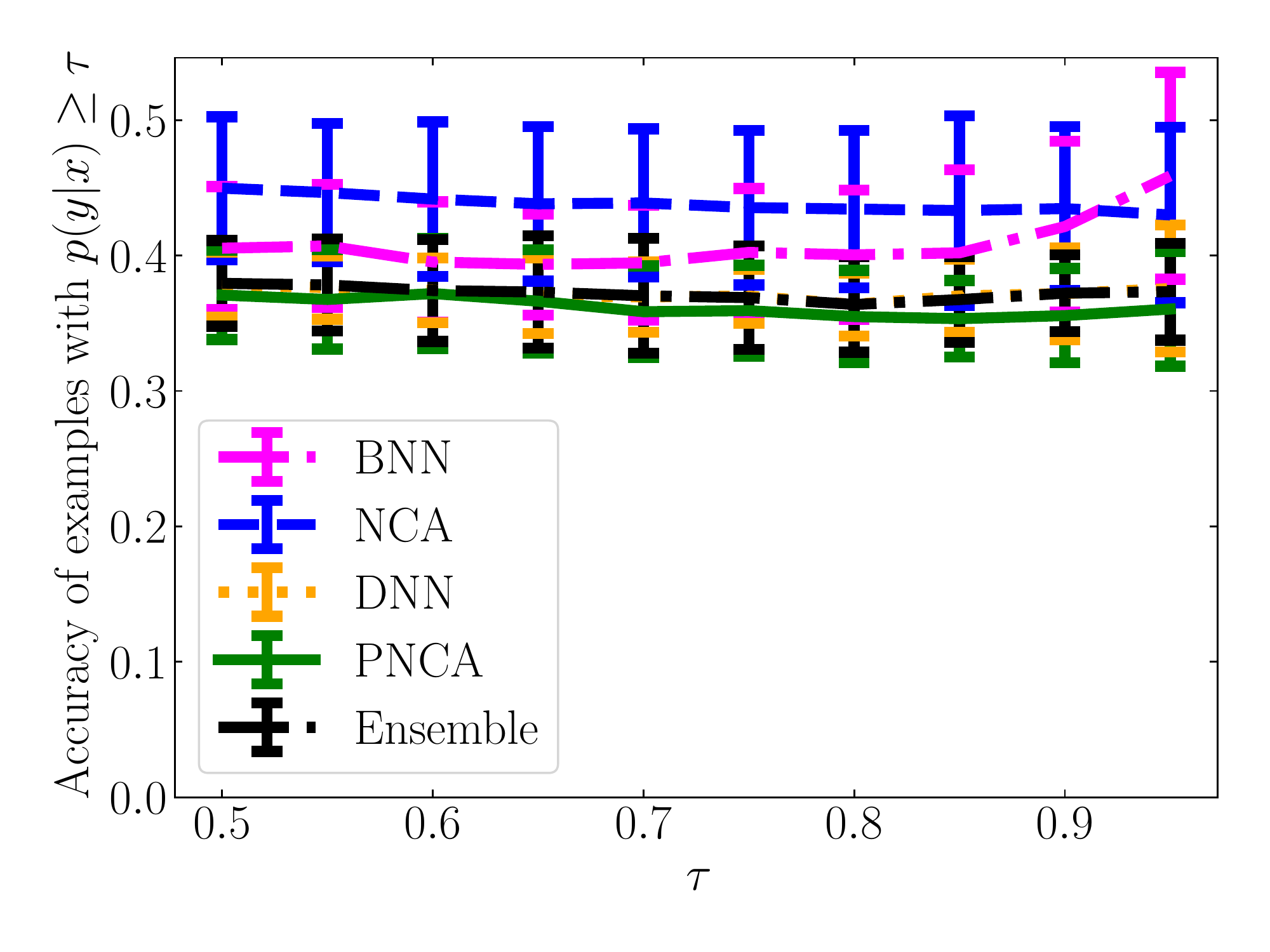}\label{fig:COVID_D2_acc}}
   \subfloat[Count (COVID-V2)]{\includegraphics[width=0.33\linewidth]{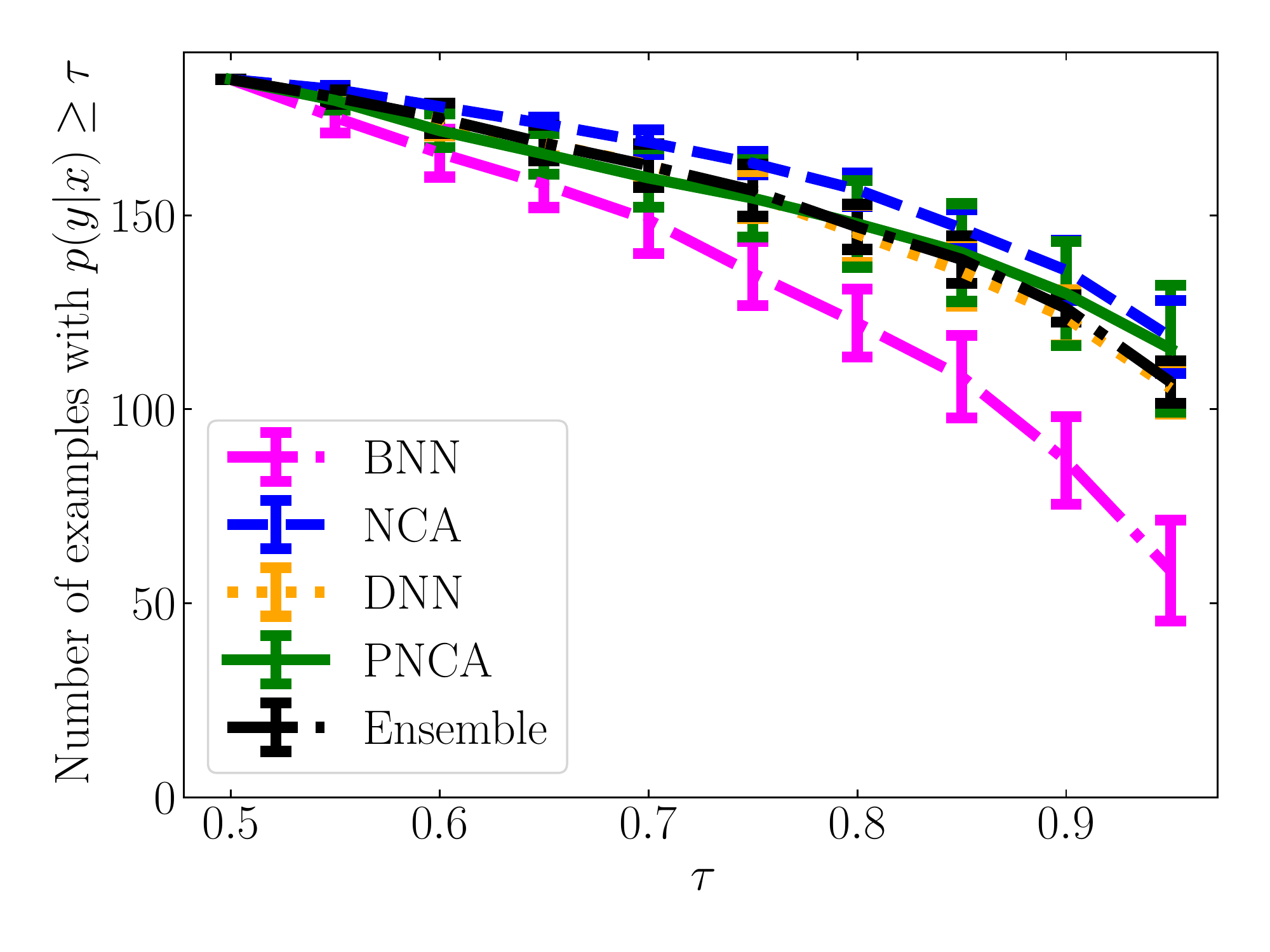}\label{fig:COVID_D2_count}}
   \subfloat[Confidence on OoD (Not-COVID)]{\includegraphics[width=0.33\linewidth]{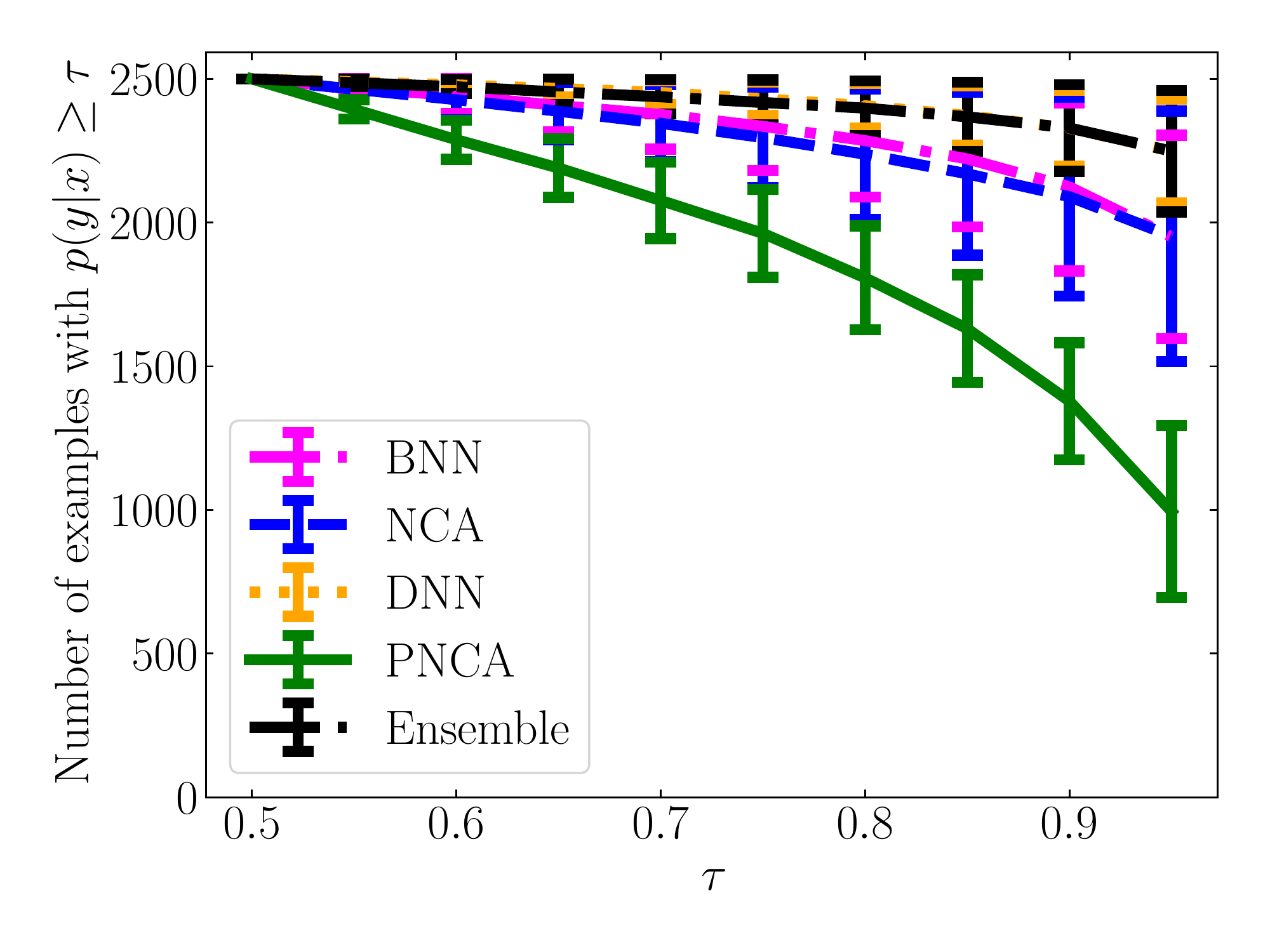}\label{fig:COVID_OOD_count}}
   \caption{While all models have similar performance on slightly shifted data (i.e., COVID-V2~\cite{ActualMed}), PNCA exhibits significantly better uncertainty quantification as the distributional shift increases (i.e., Out-of-Distribution data \cite{rajpurkar2017mura}) as seen by the much fewer high confidence predictions in (c).}
  \label{fig:COVID}
  \vspace{-2mm}
\end{figure*}

We consider two small data classification tasks: (1) handwritten digit recognition, and (2) COVID-19 detection from chest X-Ray images.
For both tasks, we compare proposed PNCA to 4 baselines, a Deep Neural Network (DNN), a Bayesian Neural Network (BNN) trained using the approach of \cite{liu2016stein}, Deep Ensembles \cite{lakshminarayanan2017simple}, and NCA~\cite{goldberger2005neighbourhood}. For PNCA and NCA, we use a neural network to map the data $\onedata$ to embeddings $\onelatent$. For NCA and PNCA, the predicted class label for a test point $\onedata_i$ is $\hat{\onetarget_i} = \argmax_\classind \sum_{j:\onetarget_j=\classind}\ncaprob_{ij}$. For the other models, the predicted class label is the one with the highest softmax probability (average softmax probability for BNNs and Ensembles). Following \cite{snoek2019can}, we use $\max_{\classind}\ncaprob(\hat{y}=\classind|\onedata)$ as a measure of the confidence of the model in predicting class $\hat{y}$ for input $\onedata$ and show the accuracy and number of examples vs. confidence for Out-of Distribution data to quantify the quality of uncertainties. Please refer to \Cref{app:Expts} for further details on the experiments and additional results.

\textbf{MNIST Classification} 
All neural networks in this experiment have the same architecture (2 hidden layers, 200 nodes per layer). 
Models are trained on a random subset of $100$ labeled examples from the MNIST dataset~\cite{lecun2010mnist} and results are averaged over $10$ trials. \Cref{fig:MNIST_rot_acc,fig:MNIST_rot_count} show the performance comparison on unseen rotated MNIST dataset (MNIST test images rotated by $60\degree$).
It can be seen that PNCA outperforms all other models in terms of (a) accuracy vs. confidence (high confidence examples should have high accuracy) and (b) number of examples with high confidence (only a few examples should have high confidence). 
Moreover, \Cref{fig:MNIST_OOD_count} shows the performance comparison on Out-of-Distribution, i.e., not-MNIST dataset~\cite{not-MNIST} that contains letters instead of handwritten digits.
We see that PNCA has significantly fewer examples with high confidence as compared to rest of the approaches on the not-MNIST dataset illustrating the superior capability in quantifying uncertainty.

\textbf{COVID-19 detection} There has been an increasing interest in using deep learning to detect COVID-19 from Chest X-Ray (CXR) images \cite{tartaglione2020unveiling,wang2020covid}. 
Successful prediction from CXR data can effectively complement the standard RT-PCR test \cite{wang2020detection}. However, the lack of large amount of training data and distributional shift between train and test data are two major challenges in this task \cite{minaee2020deep}. 

We consider two sources of COVID-19 data -- \cite{cohen2020covid}, which has been used by most existing works to train their models for COVID-19 classification and \cite{ActualMed}, which we use as our unseen test data as it comes from a different source than the images in \cite{cohen2020covid}. We follow the transfer learning approach of \cite{minaee2020deep} wherein a ResNet-50 model pre-trained on Imagenet is used as a feature extractor and the last layer of the model is re-trained on \cite{cohen2020covid} with aforementioned approaches -- DNN, BNN, Ensemble, NCA, PNCA on the training dataset. We consider a binary classification problem, i.e., each model outputs a probability $\ncaprob$ of the presence/absence of COVID-19 in a given CXR image.

We use the version of \cite{cohen2020covid} available on Kaggle\footnote{https://www.kaggle.com/bachrr/covid-chest-xray} as our training dataset, which contains $275$ COVID-19 X-Ray images and $76$ non-COVID X-Ray images. On the other hand, \cite{ActualMed} is used as our test data, which contains $58$ COVID-19 X-Ray images and $127$ non-COVID X-Ray images. There is a distributional shift present between train and test data resulting in relatively low test accuracy for all models in \Cref{fig:COVID_D2_acc}. 
We also look at the number of examples classified with a high confidence for both the test data and completely Out-of-Distribution data (shoulder and hand X-Rays from \cite{rajpurkar2017mura}). As can be seen, on \cite{ActualMed}, which potentially has a different distribution, BNN has slightly lower number of examples classified with high confidence than the other models. Next, in \Cref{fig:COVID_OOD_count}, we can see that as the distribution shift increases, PNCA makes \emph{significantly} fewer high confidence predictions than \emph{all} other models corroborating its superior uncertainty quantification. 

In summary, these experiments demonstrate that PNCA achieves much better uncertainty quantification than the baselines without losing accuracy in small-data regime. 
\section{Conclusion and Broader Impact}
This work serves as a caution to practitioners interested in applying deep learning for disease detection especially during the current pandemic since we find that the issues related to overconfident and inaccurate predictions of DNNs become even more severe in small-data regime. 
While our approach appears to be less susceptible to making overconfident misclassifications and have good uncertainty estimation performance, we acknowledge that there is still room for improvement especially with respect to the accuracy of the model. With this in mind, we will explore approaches to improve the generalization capability of PNCA in future work. Further, sample efficient uncertainty calibration approaches such as \cite{zhang2020mix} and more reliable evaluation approaches for small data regime can be explored. 

\section*{Acknowledgement}
This work was performed under the auspices of the U.S. Department of Energy by Lawrence Livermore National Laboratory under Contract DE-AC52-07NA27344 (LLNL-CONF-811603).
\bibliography{Bibliography}
\bibliographystyle{abbrv}

\appendix
\section{Additional Details on Experiments}
\label{app:Expts}
\begin{algorithm*}
\caption{PNCA}\label{algo:PNCA}
\SetKwInOut{Input}{Input}
\SetKwInOut{Output}{Output}
\Input{Training points $\alldata$ and targets $\alltarget$ along with a set of initial model parameters $\{\weights_i^{(0)}\}_{i=1}^{\numweights}$}
\Output{A set of model parameters $\{\weights_i\}_{i=1}^{\numweights} \sim \hat{\probfunc}(\weights)$ where $\hat{\probfunc}(\weights)$ is obtained by performing functional gradient descent}
Initialize model parameters $\{\weights_i^{(0)}\}_{i=1}^{\numweights} \sim \probfunc_{0}(\weights)$ for some known $\probfunc_{0}(\weights)$\\
\For{iteration $t$}{
$\weights_{i}^{(t+1)} = \weights_{i}^{(t)} - \epsilon_{t}\phi(\weights_{i}^{(t)})$,\\
where $\phi(\weights) = \sum_{l = 1}^{\numweights}\wtkernel(\weights,\weights_l)\nabla_{\weights_l}\hat{\negll}(\weights_1,\ldots,\weights_\numweights)$
}
\end{algorithm*}
\begin{figure*}[t]
  \centering
    \subfloat[Accuracy (Rotated-MNIST)]{\includegraphics[width=0.33\linewidth]{MNIST_Plots/MNIST_Rot_acc_100.pdf}}
   \subfloat[Count (Rotated-MNIST)]{\includegraphics[width=0.33\linewidth]{MNIST_Plots/MNIST_Rot_count_100.pdf}}
   \subfloat[Confidence on OoD (Not-MNIST)]{\includegraphics[width=0.33\linewidth]{MNIST_Plots/MNIST_OOD_count_100.pdf}}\\
   \subfloat[Accuracy (Rotated-MNIST)]{\includegraphics[width=0.33\linewidth]{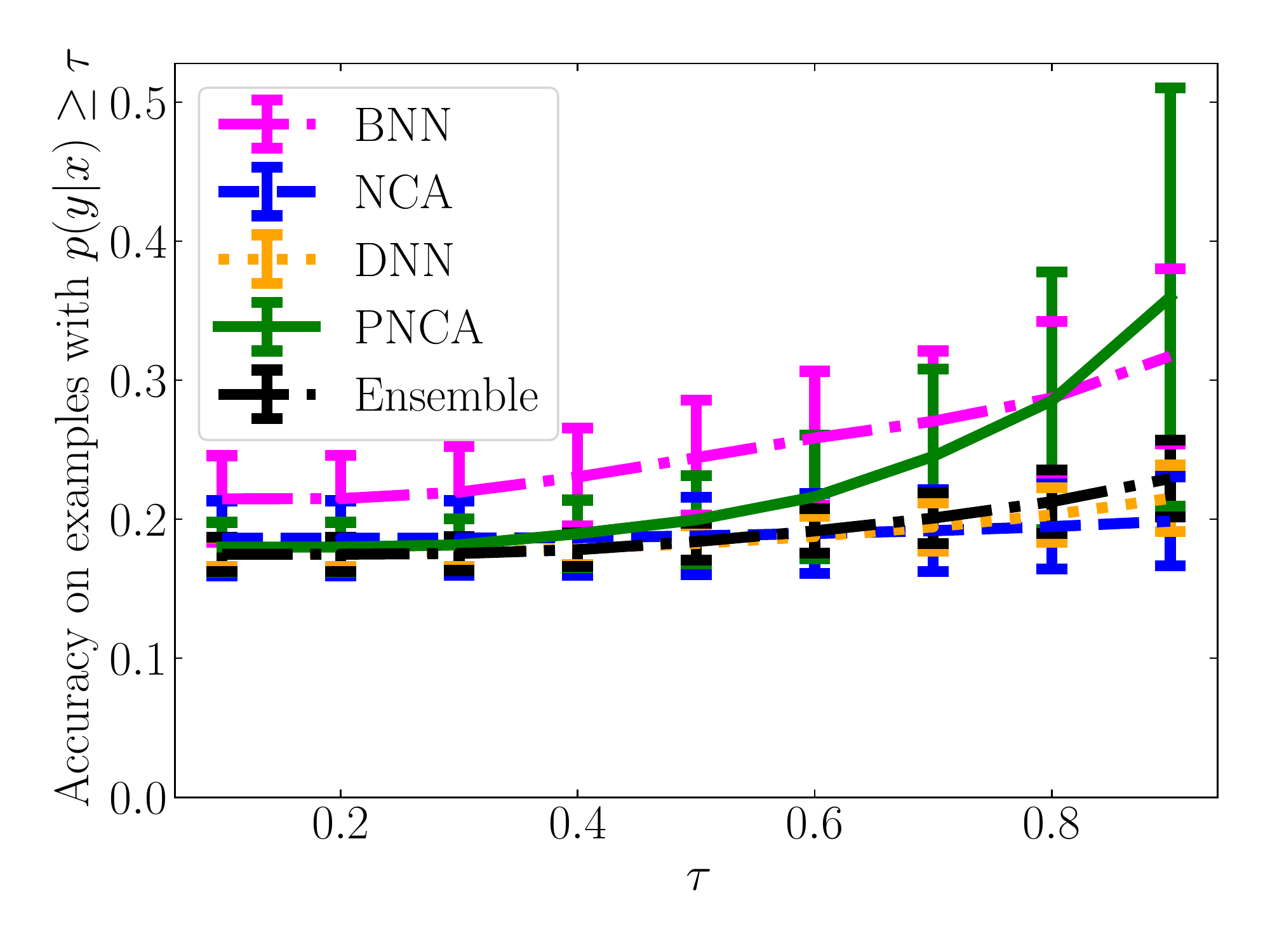}}
   \subfloat[Count (Rotated-MNIST)]{\includegraphics[width=0.33\linewidth]{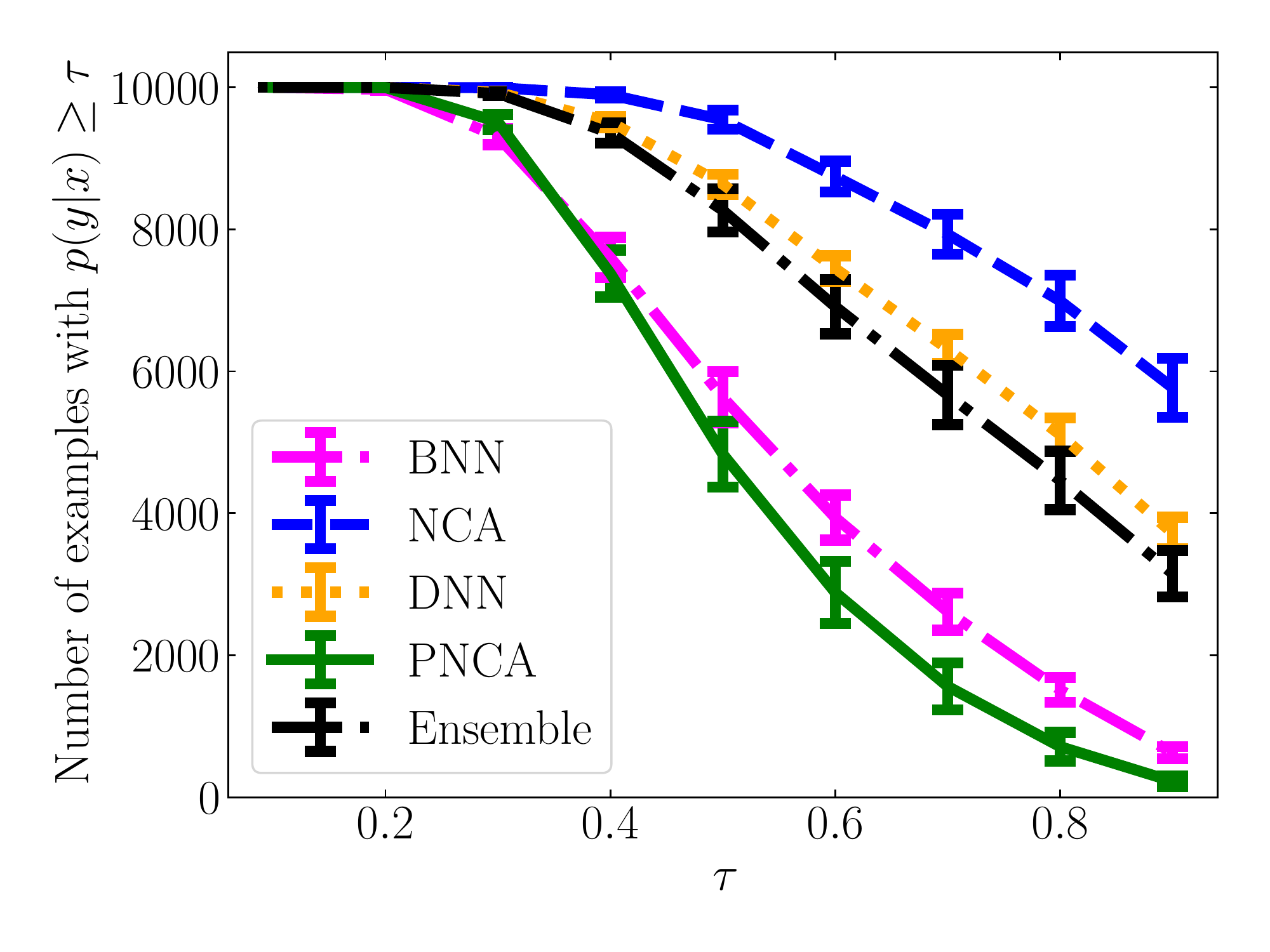}}
   \subfloat[Confidence on OoD (Not-MNIST)]{\includegraphics[width=0.33\linewidth]{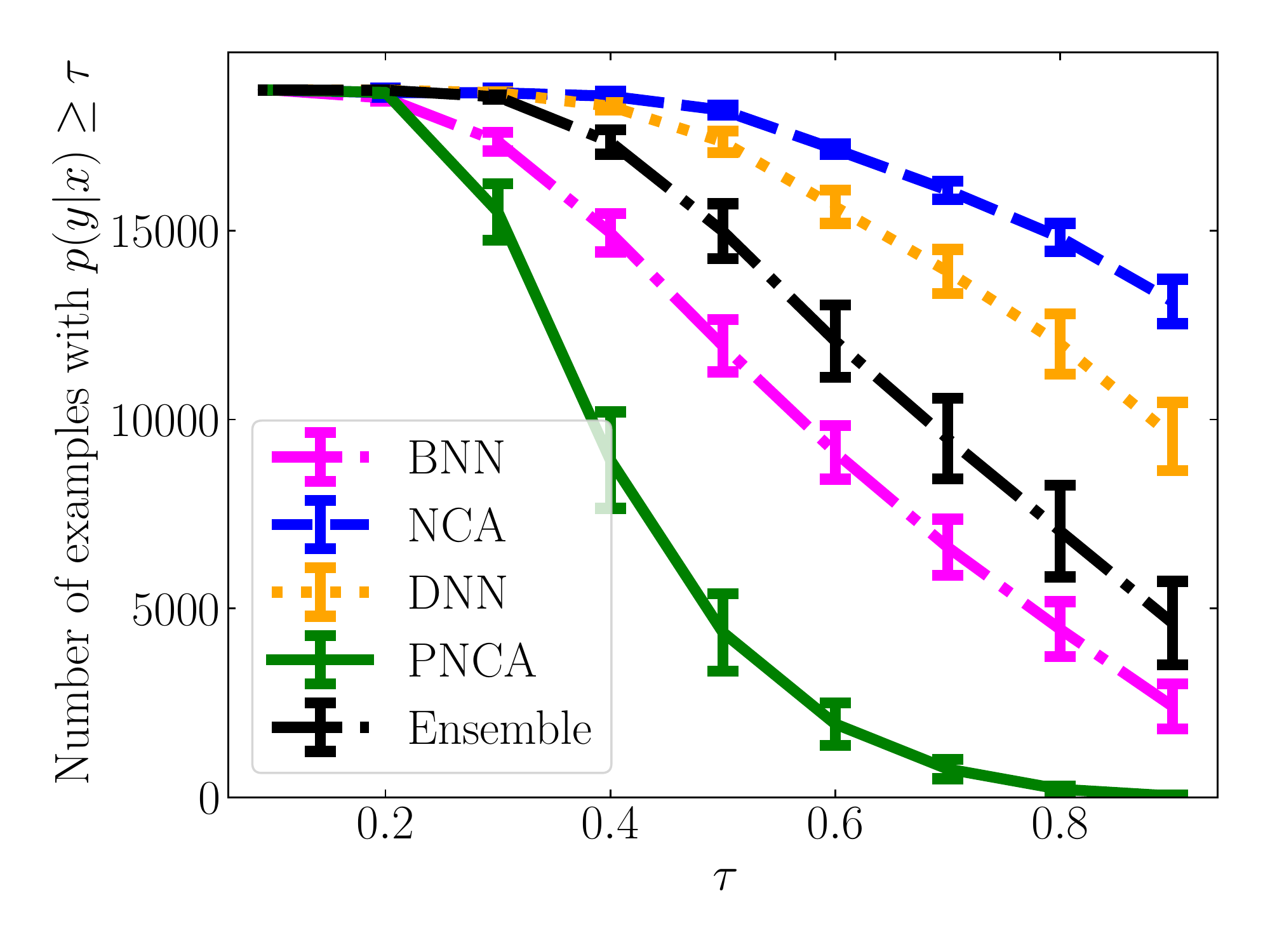}}\\
   \subfloat[Accuracy (Rotated-MNIST)]{\includegraphics[width=0.33\linewidth]{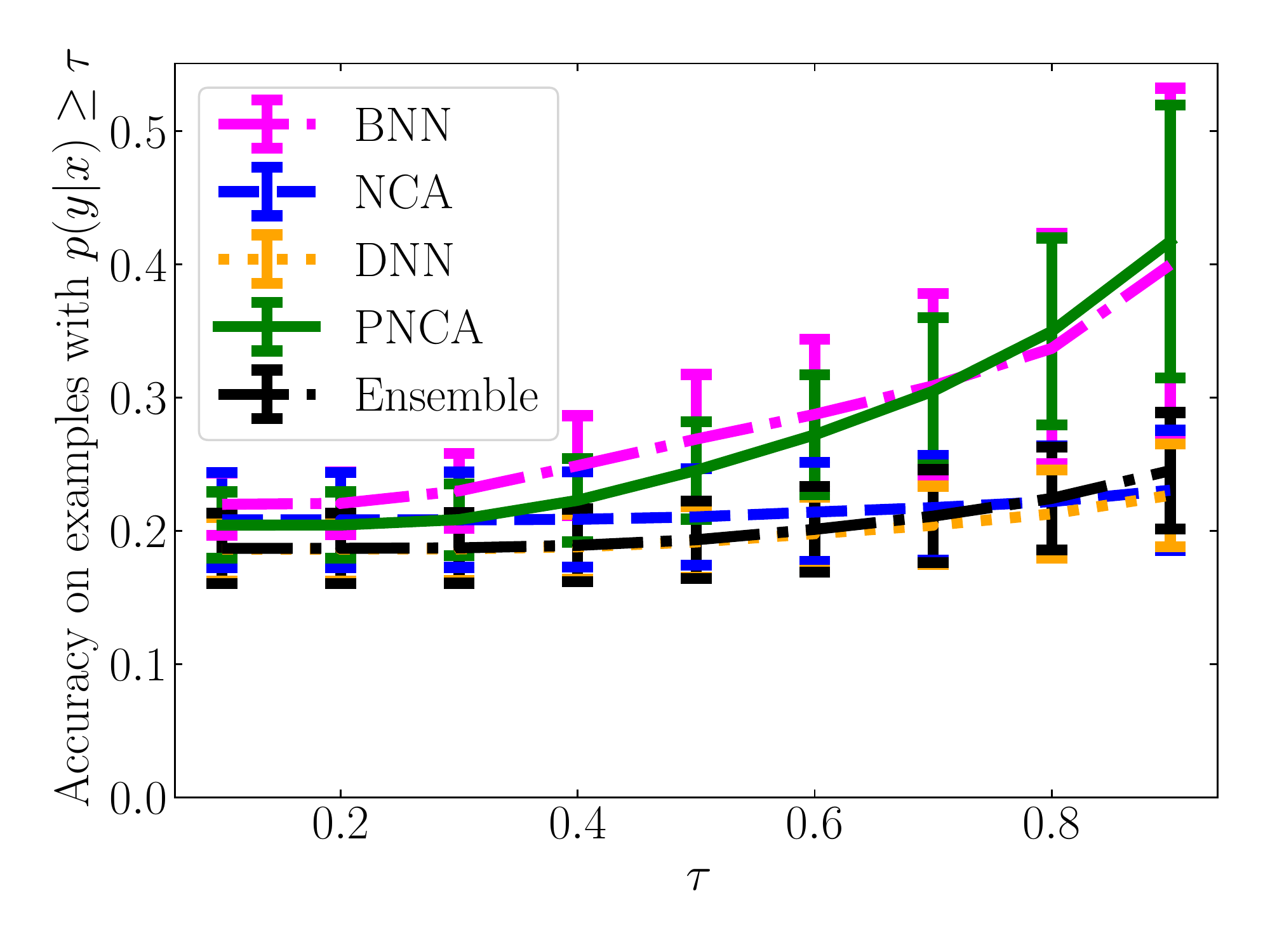}}
   \subfloat[Count (Rotated-MNIST)]{\includegraphics[width=0.33\linewidth]{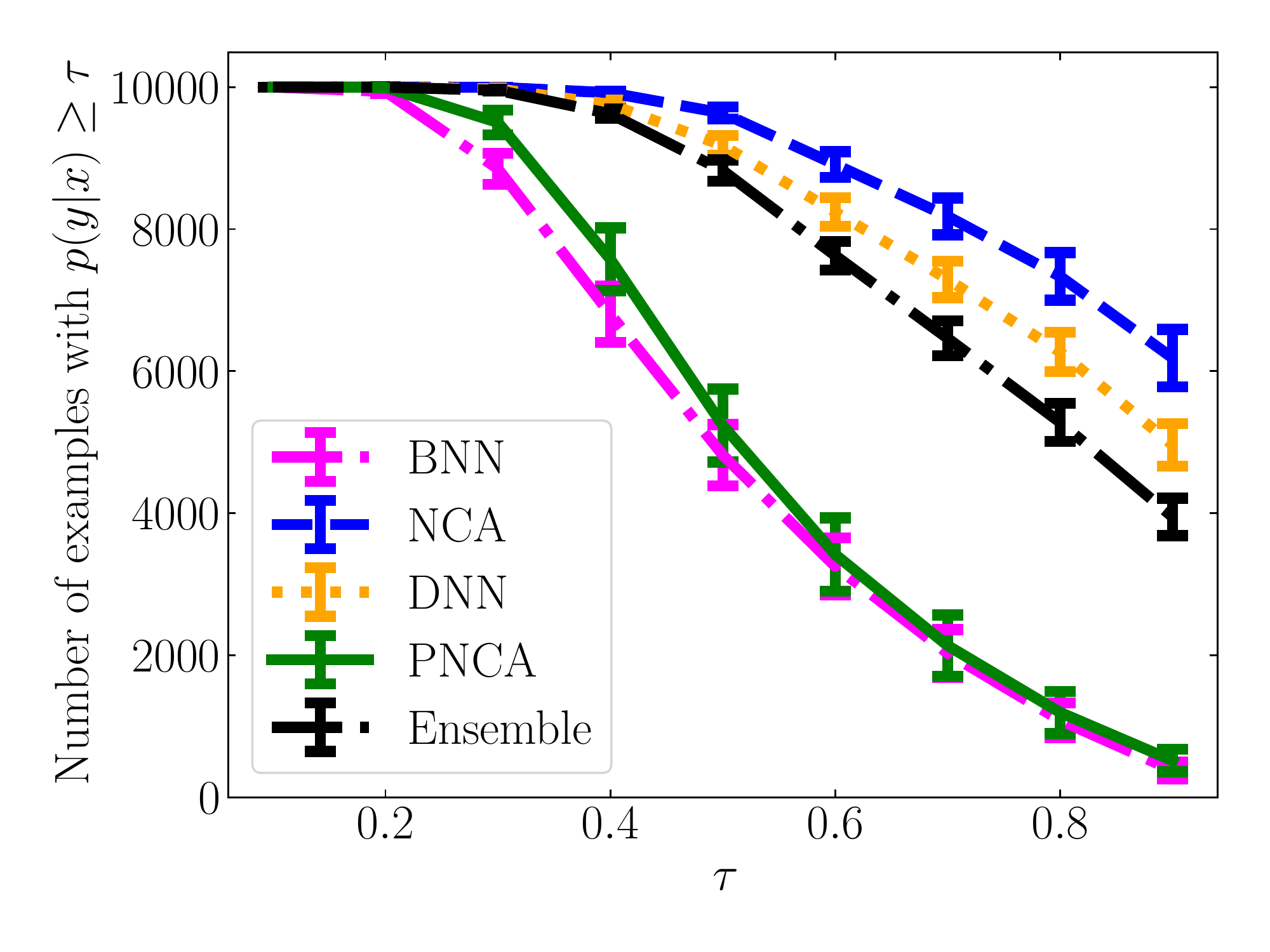}}
   \subfloat[Confidence on OoD (Not-MNIST)]{\includegraphics[width=0.33\linewidth]{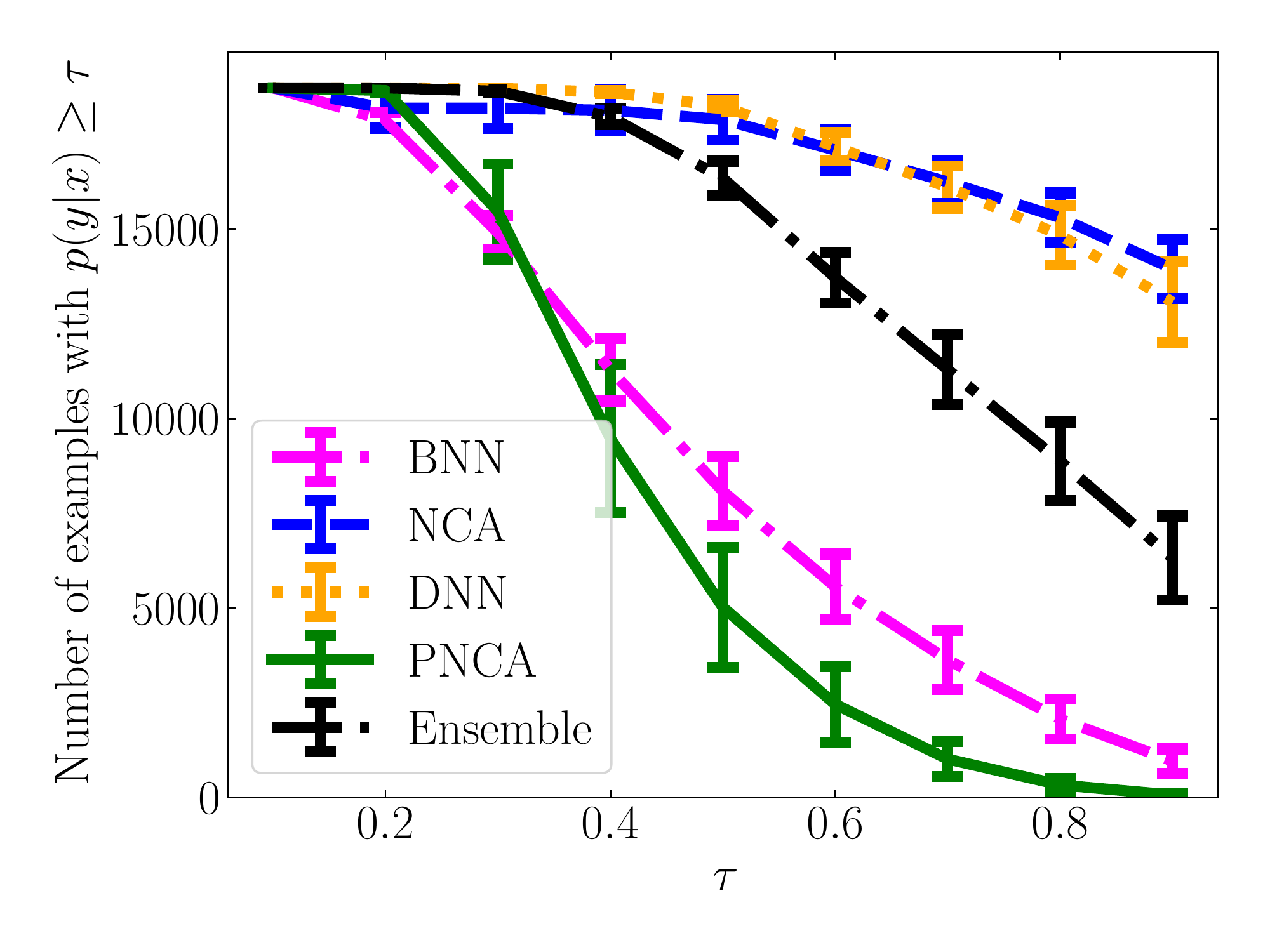}}
   \caption{Results in (a)-(c) are for $\numtrain = 100$ samples, in (d)-(f) are for $\numtrain = 200$ samples, and (g)-(i) are for $\numtrain = 400$ samples. These results show that while the performance of other models (including BNN) worsens (lower accuracy, higher confidence on OOD data) as the number of samples decreases, that of our approach PNCA is largely unaffected thus validating its efficacy in small data settings.}
  \label{fig:MNIST_app}
\end{figure*}
All models are implemented in TensorFlow \cite{abadi2016tensorflow} on a Titan X GPU with 3072 CUDA cores. We use the Adam Optimizer with Nesterov Momentum \cite{dozat2016incorporating} with a learning rate of $0.001$ to train the models for $100$ epochs. For DNN, BNN, and Ensemble we use minibatches of size $20$ with 1 epoch corresponding to a pass over the entire dataset, while for NCA and DPNCA, the entire dataset is used to calculate gradients.  

Following \cite{liu2016stein} we use the RBF kernel as the kernel $\kappa$ between model parameters $\weights$ in PNCA, with bandwidth chosen according to the median heuristic described in their work since it causes $\sum_{j}\wtkernel(\weights,\weights_j) \simeq 1$ for all $\weights$, leading $\wtkernel$ to behave like a probability distribution. We also use Orthogonal Random Features \cite{yu2016orthogonal} to approximate the kernel between probability distributions in the latent space in \cref{eq:kernij} for faster computation. We use $10\lowdim$ features where $\lowdim$ is the dimensionality of the latent space and a ReLU activation on the approximate kernel to set any spurious negative values to zero (since the original squared exponential kernel can never be negative).

\Cref{tbl:Results} contains the accuracy of different models across experiments (MNIST test data \cite{lecun2010mnist}, Rotated MNIST test data, COVID-19 validation data \cite{cohen2020covid} and COVID-19 test data \cite{ActualMed}). For MNIST, accuracy values are averaged over 10 trials (where each trial corresponds to a different set of 100 training examples). For COVID-19 accuracy values, the training data is split into 5 equal folds and in each trial we use 4 folds to train the model and the 5th fold to calculate in-distribution (validation) accuracy. Since each training data point is a part of the validation data (5th fold) only once, therefore we do not have any standard deviation values for the validation accuracy. The accuracy on COVID-19 test data is averaged across all folds.

\begin{table*}[h]
\begin{center}
\begin{tabular}{ |c | c | c | c | c| } 
 \hline
 \textbf{Method} & \textbf{MNIST Test} & \textbf{Rotated MNIST} & \textbf{COVID-19 Validation} & \textbf{COVID-19 Test} \\
 \hline
 \textbf{BNN} & $0.74 \pm 0.02$ & $0.20 \pm 0.04$ & $0.79 $ & $0.40 \pm 0.04$ \\ 
 \textbf{NCA} & $0.69 \pm 0.01$ & $0.18 \pm 0.04$ & $0.77 $ & $0.44 \pm 0.05$  \\ 
 \textbf{DNN} & $0.75 \pm 0.02$ & $0.17 \pm 0.02$ & $0.80 $ & $0.37 \pm 0.02$  \\ 
 \textbf{PNCA} & $0.67 \pm 0.03$ & $0.18 \pm 0.03$ & $0.82 $ & $0.37 \pm 0.03$ \\  
 \textbf{Ensemble} & $0.76 \pm 0.02$ & $0.17 \pm 0.03$ & $0.79$ & $0.38 \pm 0.03$  \\ 
  \hline
\end{tabular}
\end{center}
\caption{Accuracy for all Models across experiments }
\label{tbl:Results}
\vspace{-5mm}
\end{table*} 

\section{Proof of Proposition 1}
\label{app:Proof}
Observe that for any smooth one-to-one transform $\tfweights = \tf(\weights)$, $\weights \sim \probfunc(\randweights)$, the kernel between the latent distributions $\probfunc(\randlatent|\onedata_i), \probfunc(\randlatent|\onedata_j)$ corresponding to data points $\onedata_i$ and $\onedata_j$ under the transformed distribution $\probfunc_{[\tf]}(\tfweights)$ can be written as
\begin{align}
\kernel_{ij} &= \mathbb{E}_{\tfweights,\tfweights' \sim \probfunc_{[\tf]}(\tfweights)}[\smallkernel(\latfunc_{\tfweights}(\onedata_i),\latfunc_{\tfweights'}(\onedata_j))]\\
&= \mathbb{E}_{\weights,\weights' \sim \probfunc(\randweights)}[\smallkernel(\latfunc_{\tf(\weights)}(\onedata_i),\latfunc_{\tf(\weights')}(\onedata_j))]\label{eq:tfwtkernij}.
\end{align}
Since the above holds for infinitesimal shifts $\tfweights = \weights + \shift(\weights)$, a tractable choice of $\probfunc_0$ (For eg. Gaussian), enables efficient approximation of $\kernel_{ij}$ by sample averages with samples $\tfweights_i$, $\tfweights_i = \weights_i + \shift(\weights_i), \weights_i \sim \probfunc_{0}(\weights)$. 

Moreover $\kernel_{ij}$ in \Cref{eq:tfwtkernij} is a \emph{functional} of the transformation $\tf$ (for fixed $\probfunc(\randweights)$) i.e. a functional of the shift $\shift$ in our case. Therefore, the problem of finding $\probfunc^{*}(\randweights)$ in \Cref{eq:PNCA} reduces to the problem of finding the optimal shift (given $\probfunc_0(\randweights)$) i.e.
\begin{align}
    \shift^{*}(\randweights) = \argmin_{\shift(\randweights)} \negll[\shift]. \label{eq:DPKL_shift}
\end{align}
Since $\shift \in \rkhs_\wtkernel$, which is the RKHS for the kernel $\wtkernel$, we can solve \Cref{eq:DPKL_shift} via functional gradient descent.

Defining $\smallkernel_{ij}(\weights,\weights') = \smallkernel(\latfunc_{\weights}(\onedata_i),\latfunc_{\weights'}(\onedata_j))$ we have
\begin{align}
    \kernel_{ij} = \mathbb{E}_{\weights,\weights' \sim \probfunc(\randweights)}[\smallkernel_{ij}(\weights + \shift(\weights),\weights' + \shift(\weights'))]\label{eq:shiftwtkernij}
\end{align}

Assuming that the distributions $\probfunc(\randweights)$ and shifts $\shift(\randweights)$ are functions in a RKHS $\mathcal{H}$ given by the kernel $\wtkernel$ ($\wtkernel$, $\smallkernel$, and $\kernel$ are all different), we have (from the definition of functional gradient $\nabla_{\shift}\kernel_{ij}[\shift]$),
\begin{align}
    \kernel_{ij}[\shift + \epsilon\altshift] = \kernel_{ij}[\shift] + \epsilon<\nabla_{s}\kernel_{ij}[\shift],\altshift>_{\rkhs} + \mathcal{O}(\epsilon^2)
\end{align}

Thus we need to compute the difference $\kernel_{ij}[\shift + \epsilon\altshift] - \kernel_{ij}[\shift]$ which, from \Cref{eq:shiftwtkernij} is given by
\begin{align}
     \begin{split}
        \kernel_{ij}[\shift + \epsilon\altshift] - \kernel_{ij}[\shift] = \mathbb{E}_{\probfunc}[\smallkernel_{ij}(\weights + \shift(\weights) + \epsilon\altshift(\weights),\weights' + \shift(\weights') + \epsilon\altshift(\weights'))] \\
        - \mathbb{E}_\probfunc[\smallkernel_{ij}(\weights + \shift(\weights),\weights' + \shift(\weights'))]\\
    \end{split}
\end{align}

We use $\mathbb{E}_{\probfunc}$ to denote the expectation when $\weights,\weights' \sim \probfunc(\randweights)$. The above equation can be rewritten as $\kernel_{ij}[\shift + \epsilon\altshift] - \kernel_{ij}[\shift] = V_1 + V_2$ where
\begin{align}
    V_1 &= \mathbb{E}_{\probfunc}[\smallkernel_{ij}(\weights + \shift(\weights) + \epsilon\altshift(\weights),\weights' + \shift(\weights') + \epsilon\altshift(\weights'))] - \mathbb{E}_{\probfunc}[\smallkernel_{ij}(\weights + \shift(\weights) ,\weights' + \shift(\weights') + \epsilon\altshift(\weights'))]\\
    & = \epsilon\mathbb{E}_{\probfunc}[\nabla_{\weights}\smallkernel_{ij}(\weights +\shift(\weights),\weights' + \shift(\weights') + \epsilon\altshift(\weights'))\altshift(\weights))] +\mathcal{O}(\epsilon^2)\\
    \begin{split}
    &=\epsilon\mathbb{E}_{\probfunc}[(\nabla_{\weights}\smallkernel_{ij}(\weights +\shift(\weights),\weights' + \shift(\weights') + \epsilon\altshift(\weights')) \\
    &- \nabla_{\weights}\smallkernel_{ij}(\weights +\shift(\weights),\weights' + \shift(\weights')) + \nabla_{\weights}\smallkernel_{ij}(\weights +\shift(\weights),\weights' + \shift(\weights')))\altshift(\weights))] +\mathcal{O}(\epsilon^2)
    \end{split}\\
    &=\epsilon\mathbb{E}_{\probfunc}[\nabla_{\weights}\smallkernel_{ij}(\weights +\shift(\weights),\weights' + \shift(\weights')))\altshift(\weights))] + \mathcal{O}(\epsilon^2)\\
    &=\epsilon<\mathbb{E}_{\probfunc}[\nabla_{\weights}\smallkernel_{ij}(\weights +\shift(\weights),\weights' + \shift(\weights')))\wtkernel(\weights,.))],\altshift>_{\rkhs} + \mathcal{O}(\epsilon^2)
\end{align}
where the last line follows from the RKHS property.
Similarly,
\begin{align}
    V_2 &=\mathbb{E}_{\probfunc}[\smallkernel_{ij}(\weights + \shift(\weights) ,\weights' + \shift(\weights') + \epsilon\altshift(\weights'))] - \mathbb{E}_\probfunc[\smallkernel_{ij}(\weights + \shift(\weights),\weights' + \shift(\weights'))]\\
    &=\epsilon<\mathbb{E}_{\probfunc}[\nabla_{\weights'}\smallkernel_{ij}(\weights +\shift(\weights),\weights' + \shift(\weights')))\wtkernel(\weights',.))],\altshift>_{\rkhs} + \mathcal{O}(\epsilon^2)
\end{align}

Since we transform the weights $\weights$ after every iteration, therefore we only ever need to compute the gradient at $\shift(\weights) = 0$. Thus, finally, we have the expression
\begin{align}
    \nabla_{\shift}\kernel_{ij}[\shift]\mid_{\shift = 0} = \mathbb{E}_{\probfunc}[\nabla_{\weights}\smallkernel_{ij}(\weights,\weights' )\wtkernel(\weights,.) + \nabla_{\weights'}\smallkernel_{ij}(\weights,\weights')\wtkernel(\weights',.)]\label{eq:fingrad_app}
\end{align}

If we draw $\numweights$ samples of model parameters $\weights_1,\ldots,\weights_\numweights \sim \probfunc(\weights)$, the empirical estimate of $ \nabla_{\shift}\kernel_{ij}[\shift]\mid_{\shift = 0}$ given by replacing expectations with sample averages is given by
\begin{align}
    \nabla_{\shift}\kernel_{ij}[\shift]\mid_{\shift = 0} \simeq \frac{1}{\numweights^2}\sum_{l,l' = 1}^{\numweights}[\nabla_{\weights_l}\smallkernel_{ij}(\weights_{l},\weights_{l'}))\wtkernel(\weights_l,.)) + \nabla_{\weights_{l'}}\smallkernel_{ij}(\weights_{l},\weights_{l'}))\wtkernel(\weights_{l'},.))]\label{eq:empfingrad_app}
\end{align}
Without loss of generality consider all the terms in the above expression that contain the gradient with respect to $\weights_1$ and let us call that part of the summation $T_1$. Therefore
\begin{align}
    T_1 = \frac{1}{\numweights^2}\sum_{l' = 1}^{\numweights}\nabla_{\weights_1}\smallkernel_{ij}(\weights_{1},\weights_{l'}))\wtkernel(\weights_1,.)) + \frac{1}{\numweights^2}\sum_{l = 1}^{\numweights}\nabla_{\weights_1}\smallkernel_{ij}(\weights_{l},\weights_{1}))\wtkernel(\weights_1,.))\label{eq:empfingrad_w1}
\end{align}
Recall the expression for the empirical estimate of the entries of the kernel matrix $\kernel_{ij}$
\begin{align}
    \hat{\kernel}_{ij} \simeq \frac{1}{\numweights^2}\sum_{l,l' = 1}^{\numweights}\smallkernel_{ij}(\weights_{l},\weights_{l'})
\end{align}
Differentiating both sides with respect to $\weights_1$,
\begin{align}
    \nabla_{\weights_1}\hat{\kernel}_{ij} \simeq \frac{1}{\numweights^2}\sum_{l' = 1}^{\numweights}\nabla_{\weights_1}\smallkernel_{ij}(\weights_{1},\weights_{l'}) + \frac{1}{\numweights^2}\sum_{l = 1}^{\numweights}\nabla_{\weights_1}\smallkernel_{ij}(\weights_{l},\weights_{1}) \label{eq:empkerngrad_w1}
\end{align}
Note that the term $\nabla_{\weights_1}\smallkernel_{ij}(\weights_{1},\weights_{1})$ occurs in both summmations. This is because $\nabla_{\weights_1}\smallkernel_{ij}(\mathbf{u},\mathbf{v}) = \nabla_{\mathbf{u}}\smallkernel_{ij}(\mathbf{u},\mathbf{v})\nabla_{\weights_1}\mathbf{u} + \nabla_{\mathbf{v}}\smallkernel_{ij}(\mathbf{u},\mathbf{v})\nabla_{\weights_1}\mathbf{v} = \nabla_{\weights_1}\smallkernel_{ij}(\weights_{1},\weights_{1}) + \nabla_{\weights_1}\smallkernel_{ij}(\weights_{1},\weights_{1})$ when $\mathbf{u} = \mathbf{v} = \mathbf{\weights_1}$ ($\mathbf{u}, \mathbf{v}, \mathbf{\weights_1}$ are all variable)
.

Substituting \Cref{eq:empkerngrad_w1} in \Cref{eq:empfingrad_w1}
\begin{align}
    T_1 = \wtkernel(\weights_1,.)\nabla_{\weights_1}\hat{\kernel}_{ij}
\end{align}
We can apply the same argument to simplify the terms in \Cref{eq:empfingrad_app} that contain gradients with respect to other weights $\weights_2,\ldots,\weights_\numweights$ in the same fashion. Therefore,
\begin{align}
    \nabla_{\shift}\kernel_{ij}[\shift]\mid_{\shift = 0} &\simeq \frac{1}{\numweights^2}\sum_{l,l' = 1}^{\numweights}[\nabla_{\weights_l}\smallkernel_{ij}(\weights_{l},\weights_{l'}))\wtkernel(\weights_l,.)) + \nabla_{\weights_{l'}}\smallkernel_{ij}(\weights_{l},\weights_{l'}))\wtkernel(\weights_{l'},.))]\\
    &= \sum_{l = 1}^{\numweights}\wtkernel(\weights_l,.)\nabla_{\weights_l}\hat{\kernel}_{ij}
\end{align}

From the chain rule for functional gradient descent we have $\nabla_{\shift}\negll = \sum_{i,j}\frac{\partial \negll}{\partial \kernel_{ij}}\nabla_{\shift}\kernel_{ij}[\shift]$ and the corresponding empirical estimate $\nabla_{\shift}\negll\mid_{\shift = 0} \simeq \sum_{i,j}\frac{\partial \hat{\negll}}{\partial \hat{\kernel}_{ij}}\nabla_{\shift}(\sum_{l = 1}^{\numweights}\wtkernel(\weights_l,.)\nabla_{\weights_l}\hat{\kernel}_{ij})$. Switching the order of the summations gives
\begin{align}
    \nabla_{\shift}\negll\mid_{\shift = 0} \simeq \sum_{l = 1}^{\numweights}\wtkernel(\weights_l,.)\nabla_{\weights_l}\hat{\negll}(\weights_1,\ldots,\weights_\numweights)
\end{align}
\end{document}